\documentclass[runningheads]{llncs}
\usepackage[T1]{fontenc}
\usepackage{graphicx}
\usepackage{amsmath,amsfonts}
\usepackage{booktabs}
\usepackage{multirow}
\usepackage{algorithm}
\usepackage[noend]{algpseudocode}
\usepackage{url}
\usepackage[hidelinks]{hyperref}
\begin{document}
\title{Compositional Concept-Based Neuron-Level Explanations for Deep Reinforcement Learning}
\titlerunning{Compositional Concept-Based DRL Explanations}
%
\author{}
\institute{}
\author{
Zeyu Jiang\orcidID{0009-0005-7872-2155} \and
Hai Huang\thanks{Corresponding author.}\orcidID{0009-0003-7176-1018} \and
Xingquan Zuo\orcidID{0000-0001-9580-1182}
}
\authorrunning{Z. Jiang et al.}
\institute{
School of Computer Science, Beijing University of Posts and Telecommunications, Beijing 100876, China \\
\email{\{zeyujiang,hhuang,zuoxq\}@bupt.edu.cn}
}

\maketitle              
\begin{abstract}

Deep reinforcement learning (DRL) has successfully addressed many complex control problems. However, the neural networks representing policies or values remain opaque, undermining trust in high-stakes applications. While concept-based methods have shown promise in deciphering internal representations in computer vision, applying them to DRL is impeded by the absence of pre-defined semantic concepts in continuous state spaces. In this work, we propose a novel concept-based explanation framework designed to provide fine-grained, neuron-level insights into DRL models. Unlike previous approaches that rely on manual feature engineering, our framework automatically aligns neuron activations with logical formulas composed of semantic predicates. To bridge the gap between continuous signals and symbolic reasoning, we introduce a value-sensitive discretization mechanism that transforms raw state features into interpretable atomic concepts. This ensures that the vocabulary used for explanation captures strategic decision boundaries relevant to the agent's value assessment. By composing these interpretable concepts and matching them with neuron behaviors, we derive explicit explanations for the network’s internal representations. Experimental results on both continuous and discrete environments demonstrate that our method effectively identifies meaningful decision-making patterns, offering faithful explanations that align with human intuition.

\keywords{Deep Reinforcement Learning  \and Interpretability \and Explainability \and Feature Discretization}
\end{abstract}
\section{Introduction}

Deep reinforcement learning (DRL) has achieved remarkable success in solving complex sequential decision-making problems through trial-and-error learning. From game playing to robotic control, DRL has demonstrated strong capabilities across various domains. However, the increasing complexity of DRL models, particularly their neural network architectures, has created a significant challenge in comprehending their internal logic. This opacity hinders their deployment in safety-critical applications such as healthcare, autonomous driving, and financial trading, where understanding the "why" behind a decision is as crucial as the decision itself.

Existing approaches to DRL transparency primarily focus on post-hoc explanations~\cite{vouros2022explainable}. These include applying classic attribution methods like SHAP~\cite{rizzo2019reinforcement} and attention mechanisms~\cite{nikulin2019free} to identify important input features, or summarizing agent behavior through causal graphs~\cite{yu2023causal}. While these methods provide valuable insights into input-output relationships, they largely treat neural networks as black boxes, failing to reveal how individual neurons encode information and contribute to the decision-making process.

To achieve fine-grained transparency, researchers in computer vision have developed concept-based methods that match individual neurons with human-understandable concepts~\cite{bau2017network,mu2020compositional}. However, extending this paradigm to Reinforcement Learning faces a critical bottleneck: the absence of pre-defined semantic concepts. Unlike image datasets where concepts like "dog" or "texture" are inherent, RL environments often involve continuous state variables (e.g., velocity, coordinates) that do not naturally map to discrete predicates. Manually defining such concepts (e.g., "velocity $>$ 0.5") is subjective and often misaligned with the specific control logic learned by the agent.

\textbf{In this paper, we address this challenge by proposing a compositional concept-based explanation framework specifically tailored for DRL.} Our approach operates at the neuron level, decoding the internal representations of policy or value networks into explicit logical formulas. To bridge the gap between continuous signals and symbolic reasoning, we introduce a value-sensitive mechanism that transforms raw state features into interpretable atomic concepts. Instead of relying on arbitrary manual thresholds, this mechanism utilizes the trained agent's value function to guide the discretization of continuous features. By identifying intervals that correspond to significant shifts in expected return, we extract a vocabulary of interpretable concepts that are grounded in the agent's own strategic assessment.

\begin{figure}[!tb]
    \centering
    \includegraphics[width=\linewidth]{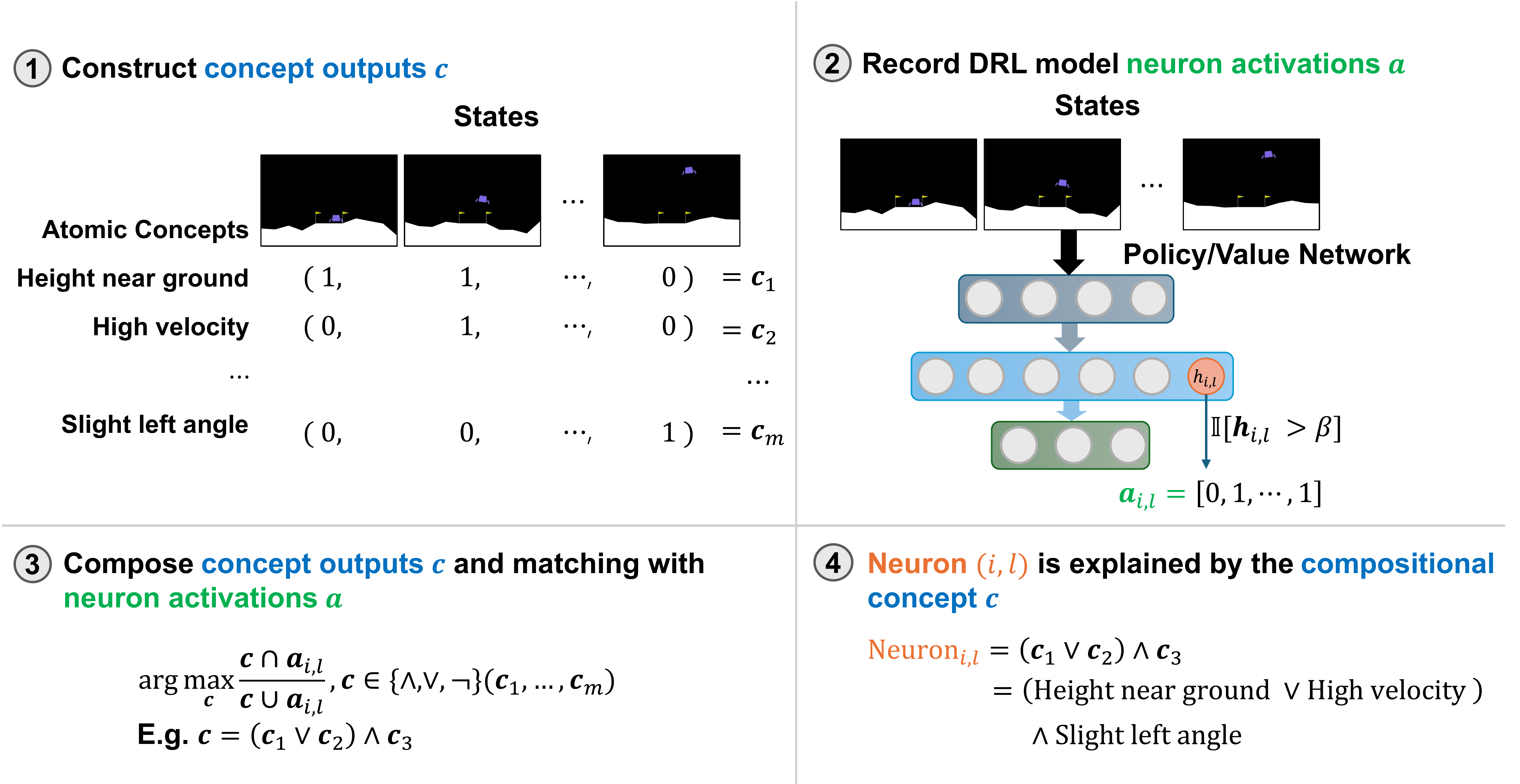}
    \caption{Our concept-based explanation framework for DRL. (1) Automatic Concept Induction: Continuous state features are partitioned into atomic concepts (e.g., "Critical Velocity") based on value function sensitivity; (2) Concept Construction: Concept vectors are generated for state sequences; (3) Neuron Analysis: DRL model activations are recorded; (4) Matching: Each neuron is explained by the compositional concept that best matches its activation pattern via beam search optimization.}
    \label{fig:overview}
\end{figure}

As illustrated in Figure~\ref{fig:overview}, our framework first induces these interpretable atomic concepts from the state space. Next, it constructs compositional formulas using logical operations. Finally, it aligns these formulas with neuron activations to generate explicit explanations for individual neurons.

To rigorously evaluate the fidelity of our explanations, we validate the framework on environments characterized by clear semantic structures and physical dynamics (Blackjack-v1, LunarLander-v3). Unlike complex tasks where the ground-truth decision logic is often obscure, these environments allow us to directly compare the derived explanations against established optimal policies. This experimental design ensures that our explanations are not only intuitive but also mathematically consistent with the agent's true behavior. Our implementation is publicly available (\href{https://github.com/WadeChiang/compositional-drl-explanations}{GitHub Repository}).

\noindent \textbf{Contributions.} The main contributions of this work are three-fold: (1) \textbf{We develop a concept-based neuron explanation framework for DRL} that reveals how individual neurons contribute to decisions using compositional logical formulas, moving beyond simple input-output attribution. (2) \textbf{We introduce a value-sensitive discretization method to generate interpretable atomic concepts.} This approach bridges the gap between raw numerical states and symbolic reasoning by automatically partitioning continuous features based on the agent's value estimates. (3) \textbf{We demonstrate the effectiveness and fidelity of our approach} on representative environments (Blackjack-v1, LunarLander-v3), showing that the framework uncovers meaningful decision-making patterns consistent with human intuition and allows for predictable behavior manipulation.

\section{Related Work}

\noindent\textbf{Explainability in reinforcement learning.} 
RL explainability is broadly categorized into self-interpretable and post-hoc methods~\cite{qing2023surveyexplainablereinforcementlearning}. Self-interpretable approaches modify architectures for transparency, employing programmatic policies~\cite{verma2018programmatically}, linear trees~\cite{liu2019lmut}, symbolic learning~\cite{landajuela2021discovering,delfosse2024interpretable,payani2020incorporating}, or concept bottlenecks~\cite{zabounidis2023concept,ye2024conceptbased}. While inherently interpretable, these often compromise expressiveness. Conversely, post-hoc methods analyze trained policies via causal modeling~\cite{yu2023causal}, language templates~\cite{boggess2023explainable,hayes2017improving}, or attribution techniques~\cite{nikulin2019free,rizzo2019reinforcement,joo2019visualization,shi2020self}. Unlike these I/O-focused methods, our work investigates the internal neural representations driving decisions.

\noindent\textbf{Concept-based explanations.} 
These methods align neural representations with human-understandable concepts. Pioneered by Network Dissection~\cite{bau2017network}, the field has evolved to include directional derivatives (TCAV)~\cite{kim2018interpretability}, automated concept discovery (ACE)~\cite{ghorbani2019towards}, and compositional explanations (CEN)~\cite{mu2020compositional}, extending to architectures like GNNs~\cite{xuanyuan2023global}. While successful in supervised learning, concept-based analysis remains underutilized in RL. We address this gap by adapting compositional methods to capture the temporal and strategic dependencies inherent in RL agents.

\section{Methodology}

\subsection{Concept Formalization}

We formalize the explanation problem by defining a mapping between the continuous state space $S \subseteq \mathbb{R}^n$ and a set of binary concepts. An \textit{atomic concept} is a function $C: S \rightarrow {0,1}$ that indicates whether a state variable lies within a specified interval (e.g., $C(s) = \mathbb{I}[y \in [0, 0.52]]$ represents "low altitude").

Compositional concepts are constructed recursively from atomic concepts using the logical operations $\mathcal{O} = \{\land, \lor, \neg\}$. For any $C_1, C_2 \in \mathcal{C}$, we define conjunction as $(C_1 \land C_2)(s) = C_1(s) \cdot C_2(s)$, disjunction as $(C_1 \lor C_2)(s) = \max(C_1(s), C_2(s))$, and negation as $(\neg C_1)(s) = 1 - C_1(s)$. A compositional concept is thus a logical formula formed by nesting these operations.

\subsection{Value-Sensitive Atomic Concept Induction}

Standard environments typically provide raw continuous features without semantic predicates. To bridge this gap, we propose an automatic induction method that discretizes continuous features based on the agent's value function $V(s)$. Given a dataset $\mathcal{D} = \{s_1, \ldots, s_N\}$ of states sampled from the 
trained agent's experience, we partition each continuous dimension into semantically 
meaningful intervals.
For actor-critic methods, we use the state value directly; for Q-learning agents, we use $V(s) = \max_a Q(s, a)$. Unlike uniform discretization, our approach is \textbf{Task-Aware}: we identify thresholds where the expected return exhibits significant variation.

For each state dimension $x_i$, we partition its domain into disjoint intervals $\{I_1, \dots, I_K\}$ via recursive partitioning that maximizes \textbf{Variance Reduction}. For a candidate threshold $\tau$ dividing dataset $\mathcal{D}$ into $\mathcal{D}_L$ (where $x_i \le \tau$) and $\mathcal{D}_R$ (where $x_i > \tau$), the variance reduction is defined as:
\begin{equation}
    \mathcal{VR}(\tau) = \text{Var}(V)_{\mathcal{D}} - \left( \frac{|\mathcal{D}_L|}{|\mathcal{D}|}\text{Var}(V)_{\mathcal{D}_L} + \frac{|\mathcal{D}_R|}{|\mathcal{D}|}\text{Var}(V)_{\mathcal{D}_R} \right).
\end{equation}
Each resulting interval $I_j$ is converted into an atomic concept $C_{i,j}(s) = \mathbb{I}[x_i(s) \in I_j]$.

\subsubsection{Determining Partition Granularity}

We select the number of intervals $K$ by minimizing reconstruction error $\mathcal{L}(K)$, defined as the MSE between $V(s)$ and the piecewise constant approximation:
\begin{equation}
    \mathcal{L}(K) = \frac{1}{N} \sum_{j=1}^{K} \sum_{s \in I_j} \left( V(s) - \bar{V}_j \right)^2,
\end{equation}
where $\bar{V}_j$ is the mean value within interval $I_j$. We use the \textit{Elbow Method}~\cite{thorndike1953belongs} to identify where marginal improvement diminishes, balancing interpretability with fidelity.

\subsection{Neural Activation Analysis}

Given the dataset $\mathcal{D}$ defined above, let $h_u(s)$ denote the activation of neuron $u$ for state $s$. Our goal is to find the concept $C \in \mathcal{C}$ that best explains when neuron $u$ activates.

\subsubsection{Similarity Measure}

We first convert continuous activations into binary indicators. For each state $s_n$, we define $a_u^{(n)} = \mathbb{I}[h_u(s_n) > \beta]$, where $\beta$ is a threshold (e.g., $\beta=0$ for ReLU networks, capturing whether the neuron is active). This yields an activation pattern $\mathbf{a}_u = (a_u^{(1)}, \ldots, a_u^{(N)}) \in \{0,1\}^N$.

Similarly, evaluating concept $C$ on each state produces a concept pattern $\mathbf{c} = (C(s_1), \ldots, C(s_N)) \in \{0,1\}^N$. We measure alignment between neuron and concept using the Jaccard similarity:
\begin{equation}
    J(\mathbf{a}_u, \mathbf{c}) = \frac{|\mathbf{a}_u \cap \mathbf{c}|}{|\mathbf{a}_u \cup \mathbf{c}|}.
\end{equation}
The explanation task is to find $C^*_u = \arg\max_{C \in \mathcal{C}} J(\mathbf{a}_u, \mathbf{c})$.

\subsubsection{Optimization via Beam Search}
We employ Beam Search (Algorithm \ref{alg:pipeline}) to explore the concept space. We maintain a beam $\mathcal{B}$ of top-$W$ candidates, expanding by applying operators from $\mathcal{O}$ and pruning lower-scoring concepts. The search runs for $D$ iterations, controlling the maximum number of composition steps. We restrict compositions to $(c_1, c_2)$ where $c_1 \in \mathcal{B}$ and $c_2 \in \mathcal{A}$, reducing complexity to $O(W \cdot |\mathcal{A}|)$ per iteration.

\begin{algorithm}[!tb]
\caption{Neuron Explanation Pipeline}
\label{alg:pipeline}
\begin{algorithmic}[1]
\State \textbf{Input:} Activations $\mathbf{a}_u$, Atomic Concepts $\mathcal{A}$, Beam Width $W$, Max Depth $D$
\State \textbf{Output:} Best Concept $C^*$
\State $\mathcal{B} \gets \mathcal{A}$, $C^* \gets \text{None}$, $\text{Score}^* \gets 0$
\For{$d = 1 \dots D$}
    \State $\mathcal{K} \gets \{ op(c_1, c_2) \mid c_1 \in \mathcal{B}, c_2 \in \mathcal{A}, op \in \{\land, \lor\} \} \cup \{ \neg c \mid c \in \mathcal{B} \}$
    \State Calculate $J(\mathbf{a}_u, \mathbf{c})$ for all $c \in \mathcal{K}$
    \State Update $C^*$ if $\max_{c \in \mathcal{K}} J > \text{Score}^*$
    \State $\mathcal{B} \gets \text{SelectTop}(\mathcal{K}, W)$
\EndFor
\State \Return $C^*$
\end{algorithmic}
\end{algorithm}

\section{Experiments}

We evaluate our method on two reinforcement learning environments: \textit{Blackjack-v1} (discrete) and \textit{LunarLander-v3} (continuous) from Gymnasium \cite{towers2024gymnasium}. Our experiments aim to answer two key questions: (1) \textbf{Explainability}: Can the proposed method automatically induce meaningful concepts and discover human-understandable neural mechanisms? (2) \textbf{Fidelity}: Are the generated explanations faithful enough to enable targeted behavior manipulation?

\subsection{Experimental Setup}

\subsubsection{Environments and Policy Training}
\begin{figure}[tb]
    \centering
    \includegraphics[width=0.8\linewidth]{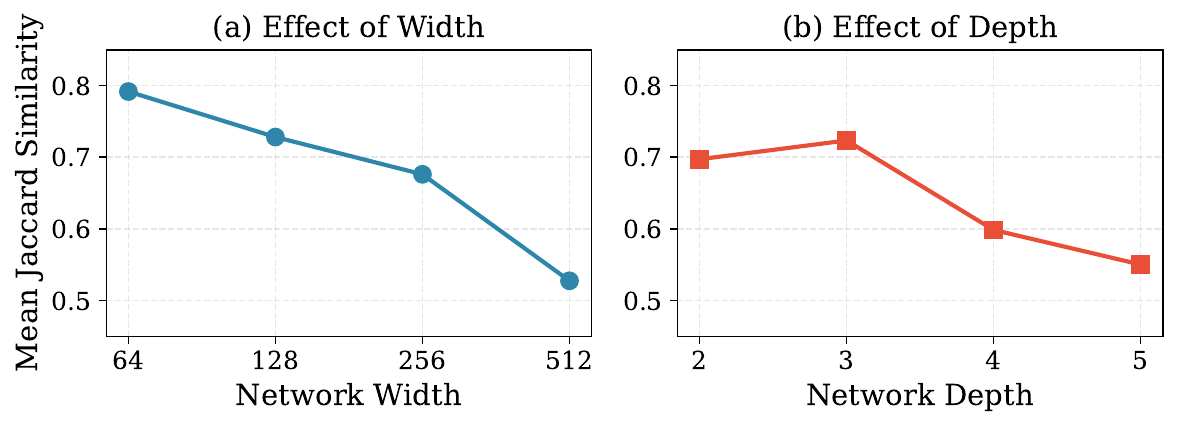}
    \caption{ Neuron interpretability (mean Jaccard Similarity) vs. network architecture. (a) Jaccard Similarity decreases monotonically with width. (b) Depth=3 achieves optimal alignment. All configurations solve the task; we select 64×3 for analysis.}
    \label{fig:arch_interp}
\end{figure}

\textit{Blackjack} features a naturally discrete 3-dimensional state space. \textit{LunarLander} involves an 8-dimensional continuous state space, comprising coordinates ($x, y$), velocities ($v_x, v_y$), angle ($\theta$), angular velocity ($\omega$), and two binary leg contact indicators.
For both tasks, we trained DQN agents with three fully connected layers (64 units each). This minimal architecture was selected based on an interpretability-capacity analysis (Figure~\ref{fig:arch_interp}): larger networks achieve similar rewards but exhibit lower neuron-concept alignment, as task knowledge becomes distributed across more units. The agents achieved near-optimal performance. We focus our analysis on the penultimate layer, as the final layer directly maps to Q-values with fixed semantics, while the penultimate layer captures the learned abstract representations.

\subsubsection{Concept Instantiation}
To generate the atomic concept set, we applied the value-sensitive partitioning described in Section 3.2. We utilized the trained agent's Q-values, $Q(s, \arg\max_a Q(s,a))$, as the target signal $y(s)$ to guide the discretization.

\begin{figure}[tb]
    \centering
    \includegraphics[width=0.8\linewidth]{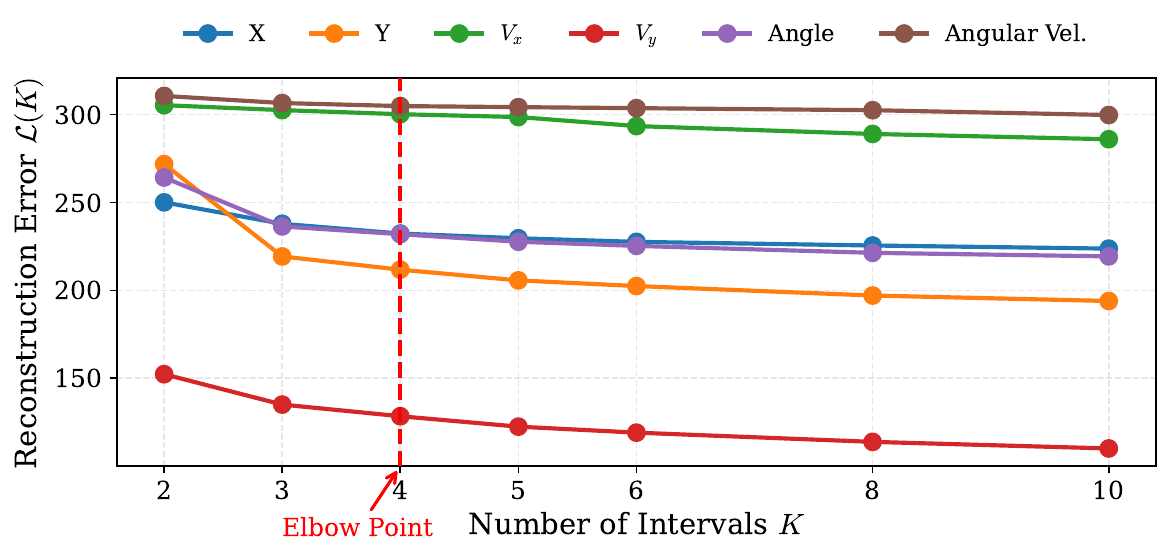}
    \caption{Reconstruction error $\mathcal{L}(K)$ versus partition granularity $K$ for each state dimension in LunarLander. The elbow point at $K=4$ balances interpretability and fidelity.}
    \label{fig:mse_analysis}
\end{figure}

\paragraph{Granularity Selection.}
We first determined the optimal number of intervals $K$ for the continuous features in \textit{LunarLander}. We computed the reconstruction error $\mathcal{L}(K)$ on a validation set of 10,000 states. As illustrated in \textbf{Fig.~\ref{fig:mse_analysis}}, the error decreases rapidly and exhibits a clear "elbow" at $K=4$. Consequently, we set $K=4$ for all continuous dimensions, balancing semantic compactness with value-approximation fidelity.

\paragraph{Induced Semantics.}
Using $K=4$, we generated the partition thresholds shown in \textbf{Table~\ref{tab:concept_thresholds}}. Each interval within a dimension constitutes an atomic concept; we denote these as $X_1, X_2, \ldots$ for horizontal position, $V_{x,1}, V_{x,2}, \ldots$ for horizontal velocity, and analogously for other dimensions, where the subscript index increases with the interval's position along the axis.

It is worth noting that although these thresholds are derived purely from value function variance, they align well with human intuition. For example, the vertical velocity $v_y$ is partitioned at $-0.48$ and $0.07$, yielding concepts that effectively distinguish semantic regimes such as $V_{y,1}$: ``fast descent,'' $V_{y,2}$: ``slow descent,'' $V_{y,3}$: ``stable hovering,'' and $V_{y,4}$: ``ascent.'' These induced intervals serve as the ground-truth vocabulary for the subsequent neuron interpretation.

\begin{table}[tb]
\centering
\caption{Automatically induced atomic concepts for LunarLander-v3 ($K=4$). The continuous state space is partitioned based on value function variance reduction. The resulting thresholds capture critical control boundaries (e.g., $v_y \approx 0$ for hovering).}
\label{tab:concept_thresholds}
\small
\setlength{\tabcolsep}{5pt}
\begin{tabular}{l c c rrr}
\toprule
\multirow{2}{*}{Feature} & \multirow{2}{*}{Range} & \multirow{2}{*}{Unit} & \multicolumn{3}{c}{Induced Thresholds} \\
\cmidrule(lr){4-6}
 & & & \multicolumn{1}{c}{$\tau_1$} & \multicolumn{1}{c}{$\tau_2$} & \multicolumn{1}{c}{$\tau_3$} \\
\midrule
Position X ($x$)   & $[-1.5, 1.5]$ & coord  & $-0.02$ & $0.12$  & $0.23$ \\
Position Y ($y$)   & $[0, 1.5]$    & coord  & $0.52$  & $0.91$  & $1.37$ \\
Velocity X ($v_x$) & $[-2.5, 2.5]$ & unit/s & $-0.61$ & $-0.34$ & $0.48$ \\
Velocity Y ($v_y$) & $[-2.5, 2.5]$ & unit/s & $-0.66$ & $-0.48$ & $0.07$ \\
Angle ($\theta$)   & $[-\pi, \pi]$ & rad    & $-0.08$ & $0.07$  & $0.13$ \\
Ang. Vel ($\omega$)& $[-5, 5]$     & rad/s  & $0.00$  & $0.004$ & $0.09$ \\
\midrule
Leg Contacts       & $\{0, 1\}$    & bool   & \multicolumn{3}{c}{\textit{N/A (Binary)}} \\
\bottomrule
\end{tabular}
\end{table}

\subsubsection{Explanation Generation Settings}
We collected a dataset $\mathcal{S}$ of 100,000 transitions for concept generation and a separate set of 10,000 states for neuron analysis. 
For the Beam Search (Algorithm 1), we configured the beam width $w=10$ and a maximum formula length of 5 to prevent overly complex explanations. Neuron activations were binarized using a threshold $\beta=0$ (ReLU cutoff). To filter out dead or redundant units, we only analyzed neurons that were active in at least 5\% of the sampled states.

\subsection{Explainability Analysis}

\subsubsection{LunarLander}
\begin{figure}[!tb]
    \centering
    \includegraphics[height=0.8\linewidth]{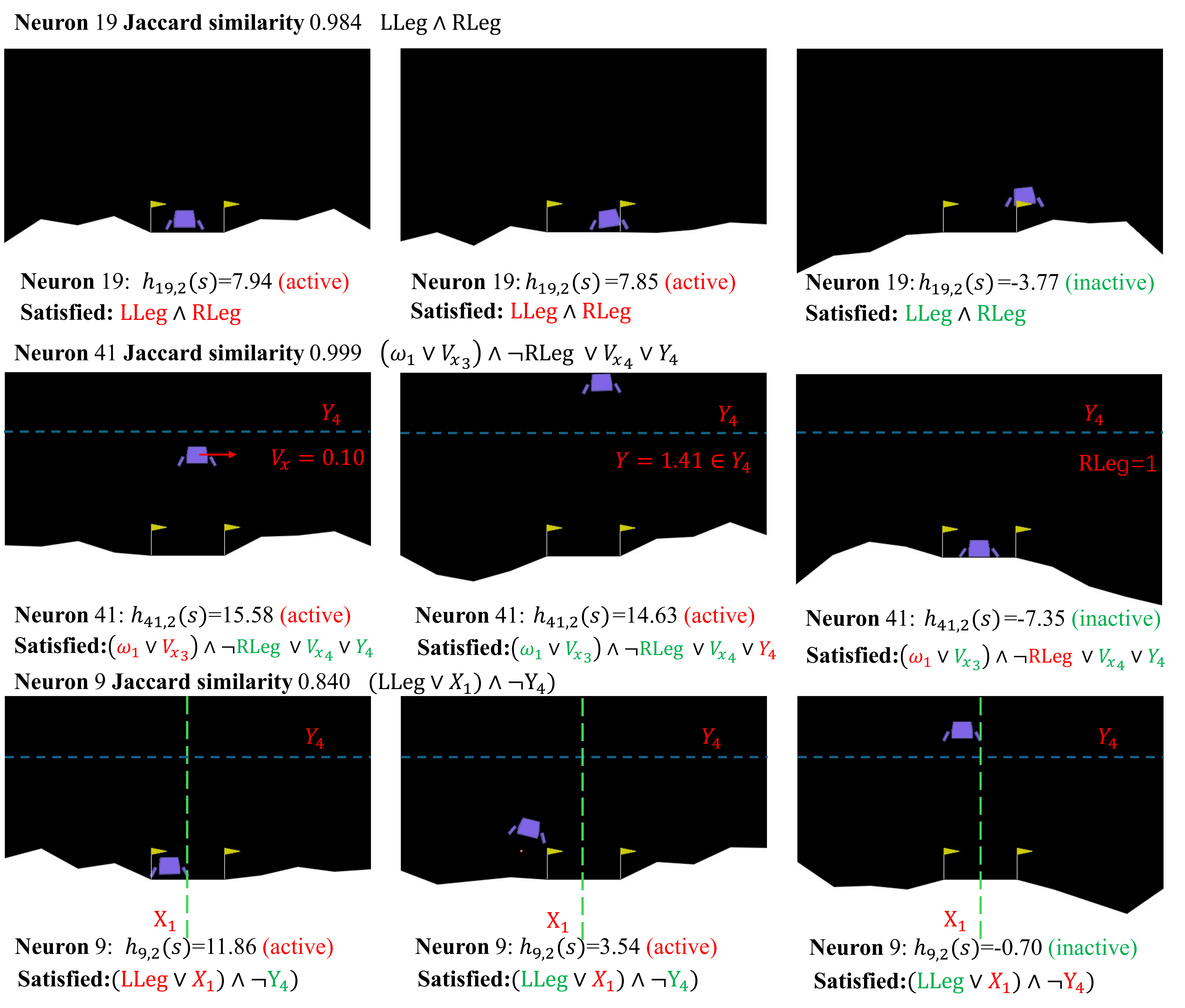}
    \caption{Visualization of three explained neurons in LunarLander. For each neuron, we display the induced logical formula, the Jaccard similarity ($J$), and three sample states (active in red, inactive in green). The results show that the neurons encode specific physical concepts: (Top) dual leg ground contact, (Middle) high-altitude or high-velocity states, and (Bottom) a specific spatial region defined by low altitude and left-side positioning.}
    \label{fig:luanrlander-discrete}
\end{figure}

In LunarLander, the induced concepts reveal structured neuron responses. As shown in Fig.~\ref{fig:luanrlander-discrete}, neurons align with distinct physical conditions.

\paragraph{Neuron 19: Ground Contact Detector.}
Neuron 19 achieves a high Jaccard similarity of $0.984$ with the formula:$\phi_{19} = \text{LLeg} \land \text{RLeg}$.
This formula indicates a strict dependency on the binary sensors for the landing gear. As shown in the top row of Fig.~\ref{fig:luanrlander-discrete}, the neuron is highly active ($h(s) \approx 7.9$) specifically when both the left and right legs are in contact with the surface. Conversely, it deactivates immediately upon liftoff. This suggests that Neuron 19 serves as a reliable indicator for the "landed" state, aggregating the discrete leg sensor inputs.

\paragraph{Neuron 41: Altitude-Velocity Composite Detector.}
Neuron 41 ($J=0.999$) encodes a composite condition involving both spatial position and kinematic dynamics. The induced formula is:$\phi_{41} = (\omega_1 \lor V_{x_3}) \land \neg \text{RLeg} \lor V_{x_4} \lor Y_4$.
    
The structure of the formula reveals that this neuron is multimodal. Specifically, the disjunction terms $Y_4$ and $V_{x_4}$ indicate that the neuron activates independently when the agent is either at a high altitude ($Y > 0.91$) \textit{or} maintaining a high horizontal velocity ($V_x \in V_{x_4}$), regardless of its vertical position. The term $(\omega_1 \lor V_{x_3}) \land \neg \text{RLeg}$ further captures states where the lander is airborne with specific rotational or moderate lateral movements. Consequently, Neuron 41 does not track a single physical variable but rather monitors a "high-energy" state characterized by either significant potential energy (altitude) or kinetic energy (velocity).

\paragraph{Neuron 9: Low-Altitude Left Region Monitor.}
Neuron 9 ($J=0.840$) encodes a spatial condition via 
$\phi_{9} = (\text{LLeg} \lor X_1) \land \neg Y_4$, 
where $X_1$ denotes the far-left region ($x < -0.02$) 
and $\neg Y_4$ indicates non-peak altitude ($y \le 0.91$). 
As shown in Fig.~\ref{fig:luanrlander-discrete}, the neuron 
activates when the lander occupies the lower-left quadrant, 
with \text{LLeg} extending activation to ground-contact states 
in this region. This demonstrates that the agent's internal 
representation partitions the state space into distinct 
spatial zones.

\subsubsection{Blackjack}
In Blackjack, the interpreted neurons (Table \ref{tab:blackjack_neurons}) align closely with standard card counting strategies.
Neuron 28 identifies "safe" hands ($17 \le \text{Sum} \le 21$), strongly contributing to the \textit{Stick} action ($w_{stick} > w_{hit}$). Conversely, Neuron 13 detects "weak" hands ($6 \le \text{Sum} \le 10$), driving the agent to \textit{Hit}.
More subtly, Neuron 60 captures a risk condition: holding a high sum (20-21) without a usable Ace against a Dealer's 10. Its negative weights for both actions suggest it acts as a value suppressor in almost-lost situations. These results show the method can extract exact symbolic rules from the neural policy.

\begin{table}[tb]
\centering
\caption{Explained neurons in Blackjack. Concepts align with game rules, and weights ($w$) indicate their contribution to action selection.}
\label{tab:blackjack_neurons}
\small
\begin{tabular}{r c l r r}
\toprule
ID & Jacc. & Concept & $w_{stick}$ & $w_{hit}$ \\
\midrule
28 & 0.92 & $P_{17} \lor P_{18} \lor P_{19} \lor P_{20} \lor P_{21}$ & $\mathbf{0.37}$ & $0.16$ \\
13 & 0.79 & $P_6 \lor P_7 \lor P_8 \lor P_9 \lor P_{10}$ & $-0.81$ & $\mathbf{-0.35}$ \\
17 & 0.89 & $D_7 \lor D_8 \lor D_9 \lor D_{10}$ & $-0.20$ & $-0.08$ \\
6 & 0.89 & $\neg P_9 \land \neg P_{10} \land \neg P_{11}$ & $0.08$ & $-0.19$ \\
60 & 0.60 & {\small $\text{NoAce} \wedge P_{20} \vee P_{21} \wedge D_{10}$} & $-0.148$ & $-0.268$ \\
\bottomrule
\end{tabular}
\end{table}

\subsection{Validation via Targeted Perturbation}

To verify that our explanations causally explain the agent's behavior, we conduct targeted perturbation experiments. We identify critical neurons for specific actions and modify the state variables involved in their assigned concepts to observe changes in activation and decision-making.

\begin{figure}[!tb]
    \centering
    \includegraphics[width=\linewidth]{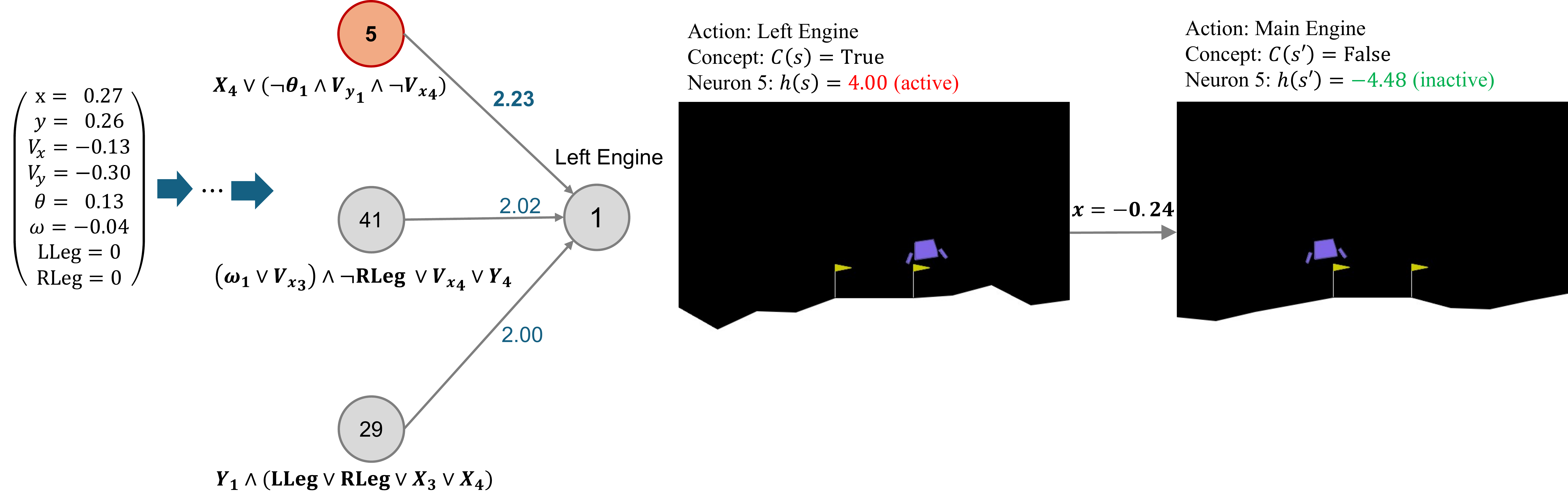}
    \caption{Causal validation via targeted perturbation in LunarLander. 
    Neuron 5, the primary contributor to the \textit{Left Engine} action, 
    encodes concept $X_4$ (right-side spatial region). 
    (Left) Original state ($x=0.27$): concept holds, $h=4.00$, agent selects \textit{Left Engine}. 
    (Right) After perturbation ($x=-0.24$): concept violated, $h=-4.48$, 
    agent switches to \textit{Main Engine}.}
    \label{fig:lunar_perturbation}
\end{figure}

\paragraph{LunarLander Case Study.}
We analyze the decision logic for Action 1 (\textit{Left Engine}) 
via its top-3 contributing neurons (5, 41, and 29), focusing on 
Neuron 5, whose formula is dominated by the spatial term $X_4$, 
representing the agent being positioned to the right of the landing pad.
As shown in Figure~\ref{fig:lunar_perturbation}, when this concept 
is satisfied in the original state, Neuron 5 activates strongly and 
the agent selects the \textit{Left Engine} to correct its trajectory. 
A targeted perturbation shifting the lander to the left violates 
the concept, deactivates Neuron 5, and causes the agent to switch 
to the \textit{Main Engine}. This confirms that Neuron 5 functions 
as a spatial trigger explicitly encoding the ``correction from right'' maneuver.

\paragraph{Blackjack Case Study.}
We extend this validation to Blackjack (Table \ref{tab:blackjack_perturbation}). For Neuron 28 (High Sum), perturbing the player's hand from 20 to 14 deactivates the neuron and flips the action from \textit{Stick} to \textit{Hit}. Similarly, for Neuron 17 (Dealer Strong), changing the dealer's card from 9 to 5 deactivates the risk-sensing neuron.
In all cases, manipulating the features defined by our concepts leads to predictable behavioral changes, demonstrating that our method extracts the true decision logic of the Deep RL agent.

\begin{table}[!tb]
\centering
\caption{Perturbation results in Blackjack. Violating the concept condition consistently flips neuron activation (Active $\to$ Inactive) and alters the agent's decision.}
\label{tab:blackjack_perturbation}
\scriptsize 
\begin{tabular}{@{}r l l l@{}}
\toprule
ID & Concept & Original State $\to$ Action & Perturbed $\to$ Action \\
\midrule
28 & $P_{\ge 17}$ & (20, 9, 0) $\to$ \textbf{Stick} & (14, 9, 0) $\to$ \textbf{Hit} \\
   & & $h=2.04$ (Active) & $h=-1.03$ (Inactive) \\
\midrule
13 & $P_{\le 10}$ & (6, 9, 0) $\to$ \textbf{Hit} & (17, 9, 0) $\to$ \textbf{Stick} \\
   & & $h=2.04$ (Active) & $h=-6.89$ (Inactive) \\
\midrule
17 & $D_{\ge 7}$ & (15, 9, 0) $\to$ \textbf{Hit} & (15, 5, 0) $\to$ \textbf{Stick} \\
   & & $h=1.07$ (Active) & $h=-0.92$ (Inactive) \\
\bottomrule
\end{tabular}
\end{table}

\section{Conclusion}

We presented a framework for extracting neuron-level explanations 
in Deep Reinforcement Learning as human-readable logical formulas, 
supported by Value-Sensitive Atomic Concept Induction that 
automatically derives semantically meaningful state discretizations 
without manual annotation. Experiments on Blackjack-v1 and 
LunarLander-v3 demonstrate interpretable neuron functions aligned 
with domain semantics, with perturbation-based validation confirming 
their causal rather than correlational nature. Future work will 
extend this to high-dimensional visual inputs and formal safety 
verification in autonomous systems.
%
%
\bibliographystyle{splncs04}
\bibliography{pakdd}

\end{document}